%% file: main.tex
\documentclass[10pt,twocolumn,letterpaper]{article}

\usepackage{cvpr}              


\definecolor{cvprblue}{rgb}{0.21,0.49,0.74}
\usepackage{multirow,tabularx,makecell,amssymb}
\usepackage[pagebackref,breaklinks,colorlinks,allcolors=cvprblue]{hyperref}

\def\paperID{8336} 
\def\confName{CVPR}
\def\confYear{2025}

\title{FIRE: Robust Detection of Diffusion-Generated Images via Frequency-Guided Reconstruction Error}

\author{Beilin Chu, Xuan Xu, Xin Wang, Yufei Zhang, Weike You, Linna Zhou\\
School of CyberSpace Security, Beijing University of Posts and Telecommunications\\
{\tt\small beilin.chu@bupt.edu.cn}
}
\begin{document}
\maketitle
\input{sec/0_abstract} 
\input{sec/1_intro}
\input{sec/2_related_works}
\input{sec/3_method}
\input{sec/4_experiment}
\input{sec/5_conclusion}
{
    \small
    \bibliographystyle{ieeenat_fullname}
    \bibliography{main}
}

\input{sec/X_suppl}

\end{document}

%% file: sec/0_abstract.tex
\begin{abstract}
    The rapid advancement of diffusion models has significantly improved high-quality image generation, making generated content increasingly challenging to distinguish from real images and raising concerns about potential misuse. In this paper, we observe that diffusion models struggle to accurately reconstruct mid-band frequency information in real images, suggesting the limitation could serve as a cue for detecting diffusion model generated images. Motivated by this observation, we propose a novel method called \textbf{F}requency-gu\textbf{I}ded \textbf{R}econstruction \textbf{E}rror (FIRE), which, to the best of our knowledge, is the first to investigate the influence of frequency decomposition on reconstruction error. FIRE assesses the variation in reconstruction error before and after the frequency decomposition, offering a robust method for identifying diffusion model generated images. Extensive experiments show that FIRE generalizes effectively to unseen diffusion models and maintains robustness against diverse perturbations. Our code is available at \url{https://github.com/Chuchad/FIRE}.
\end{abstract}

%% file: sec/1_intro.tex
\section{Introduction}
\label{sec:intro}

The advent of diffusion models (DMs) \cite{ho2020denoising} has made high-quality, controllable image generation significantly more accessible. Subsequent works leverage large-scale multimodal learning \cite{ruiz2023dreambooth,rombach2022high,shi2024instantbooth}, particularly involving text and image pairs, to further diversify the scenarios in which DMs can generate images. The realism of these generated images has reached a level where they can deceive the human visual system, raising concerns about the potential misuse of DMs to create misleading content \cite{juefei2022countering}. Therefore, the development of effective methods to detect such generated images is critical to mitigate the potential societal risks posed by this technology.

Several works \cite{baraldi2024contrasting,zhang2023diffusion,he2024rigid,zhong2024patchcraft} have been proposed specifically for detecting images generated by DMs. Among them, a novel class of detection methods based on diffusion reconstruction are proposed \cite{wang2023dire,ricker2024aeroblade,ma2023exposing,luo2024lare,cazenavette2024fakeinversion}, highlighting a promising research direction. These methods are founded on the assumption that generated images, having undergone the diffusion process, are closer to the latent space in DM. Consequently, they are less affected by reconstruction through a second denoising process compared to real images. By utilizing the reconstruction error, these methods can achieve satisfactory performance in detecting images generated by various models, often with lightweight models or even in a training-free manner. However, these methods still face challenges in generalizing to unseen datasets \cite{cazenavette2024fakeinversion}.

\begin{figure}
    \centering
    \includegraphics[width=\linewidth]{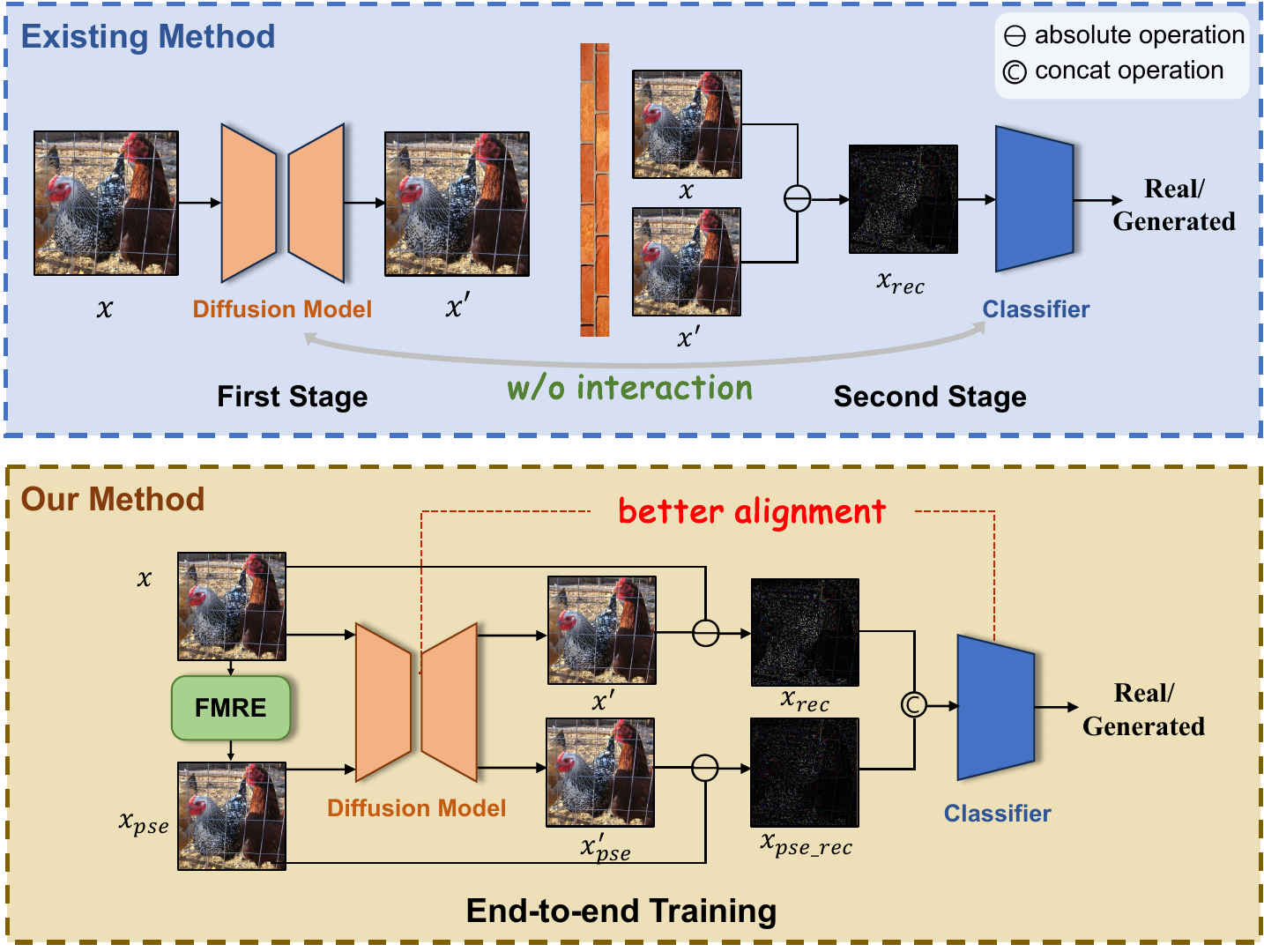}
    \caption{Comparison between existing reconstruction-based methods and FIRE. Existing approaches \cite{wang2023dire,ricker2024aeroblade} proceed in two steps: first, compute the reconstruction error of the image using a pre-trained diffusion model, and then train a backend classifier on the reconstruction error. FIRE integrates the classifier with the diffusion model, allowing \textbf{end-to-end} learning and better alignment of the latent space for artifact generation and detection. Additionally, FMRE can leverage \textbf{frequency-guided reconstruction} to identify the information that the diffusion model struggles to reconstruct.}
    \label{fig:fi0}
\end{figure}

\begin{figure*}[ht]
    \centering
    \includegraphics[width=0.8\linewidth]{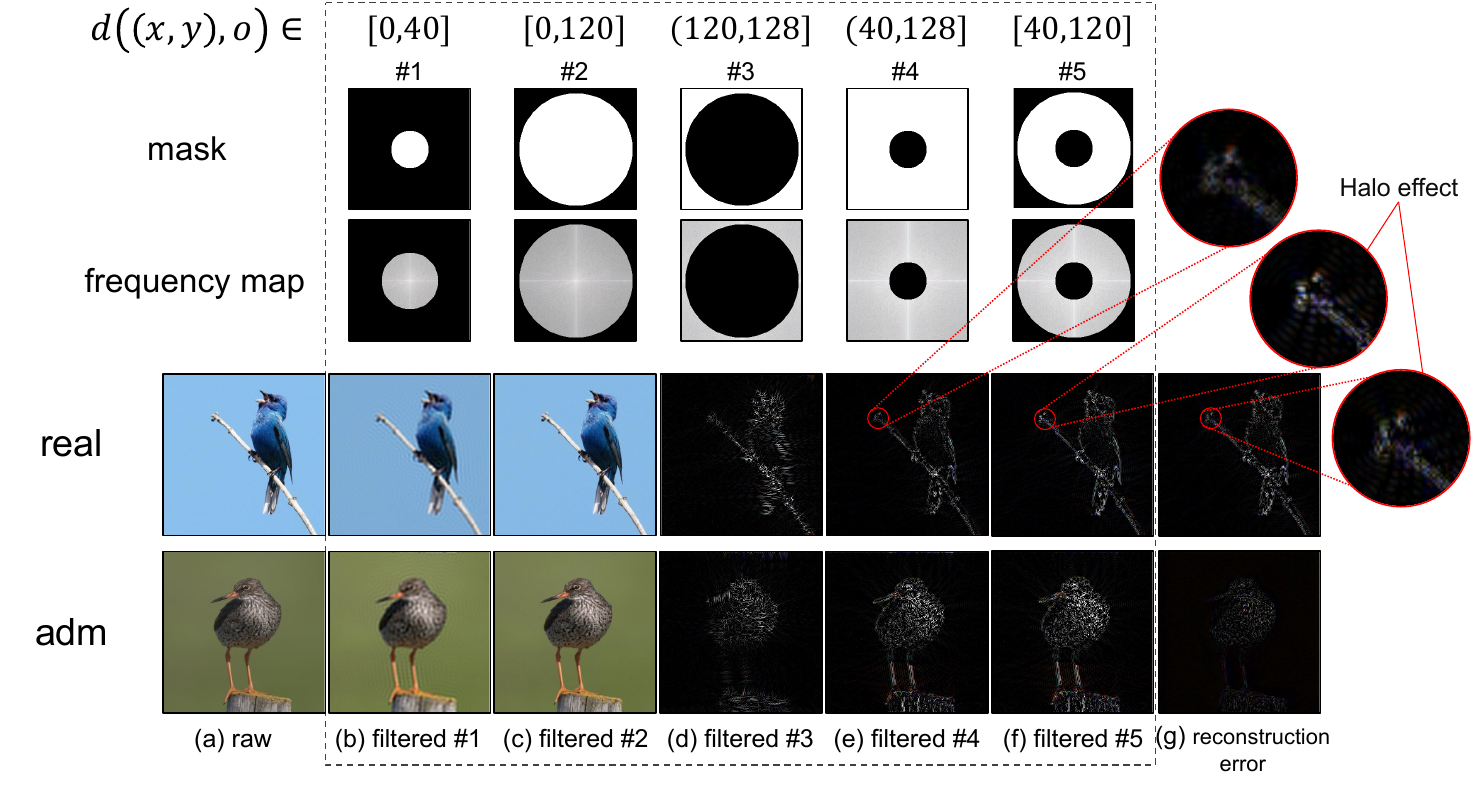}
    \caption{\textbf{Analysis between results of images filtered by different masks and their reconstruction errors.} (a) shows a real image from ImageNet and a generated counterpart produced by a pre-trained ADM \cite{dhariwal2021diffusion} model. (b)-(f) are the results of applying different frequency masks to the image. Frequency maps are obtained by applying FFT to the images, with the low-frequency regions shifted to the center. $d((x,y),o) \in$ represents points within a specific distance range from the center of the mask, which are set to 1 (preserved), while the rest are filtered out (set to 0). (g) shows the reconstruction error using Stable Diffusion v1.5 \cite{runwayml2023sd}. The red circles highlight the halo effect observed in \textbf{filtered \#5} and \textbf{the reconstruction error}. The reconstruction error of the real image visually resembles the band-pass filtered~\#5 image. Notably, as explained in prior work \cite{wang2023dire}, the reconstruction error for generated images is much lower than for real ones. (To enhance print visibility, we apply 100\% sharpening to the residual images in all figures presented in the main paper.)}
    \label{fig:fig1}
\end{figure*}

When rethinking the significance of reconstruction error representation in the diffusion process, we argue it indeed reflects the parts of the image that the DM struggles to accurately reconstruct. Therefore, existing reconstruction-based methods can be viewed as a process of identifying which parts of the image that are challenging to reconstruct. However, current approaches input the entire reconstruction error into a backend classifier, which also contains both potential content biases and noise from the diffusion process. This might be the reason why existing methods struggle to generalize on unseen datasets \cite{cazenavette2024fakeinversion}. Luo et al. \cite{luo2024lare} observe that the reconstruction error predominantly manifests in the high-frequency components of the image, yet their approach still lacks the refinement of the hard-to-reconstruct information. Therefore, a critical research questions arise: which specific parts of the image are particularly challenging for the DM to reconstruct?

To address the question, we conduct a frequency analysis of real and generated images. As shown in Figure~\ref{fig:fig1}, low-frequency information captures more of the color and general content, while noise and edge details are embedded in the high-frequency components. The mid-frequency band (filtered \#5 in Figure~\ref{fig:fig1}), which contains both high- and low-frequency information, visually aligns well with the patterns seen in the reconstruction error maps. Based on this observation, \textbf{we hypothesize that the reconstruction error—i.e., the hard -to-reconstruct information, primarily resides in the mid-frequency region.} Visualization experiments in Section~\ref{sec:vision} corroborate this hypothesis. Moreover, the mid-frequency component and reconstruction error of real images exhibit stronger signals than generated images. This suggests that real images contain unique mid-frequency information that is particularly difficult to reconstruct. Intuitively, if we can isolate this mid-frequency signal from the image, comparing the reconstruction errors before and after this isolation could serve as a potential cue for detecting generated images.
\label{sec:intro_hypothesis}

Building on the above observation and hypothese, we propose a novel \textbf{F}requency-gu\textbf{I}ded \textbf{R}econstruction \textbf{E}rror (\textbf{FIRE}) method for detecting DM-generated images. FIRE consists of a \textbf{F}requency \textbf{M}ask \textbf{RE}finement module (FMRE) and a backend classifier. FMRE refines the localization of the mid-frequency regions that are difficult for the DM to reconstruct, and filters out such frequency component. The backend classifier then evaluates the change in reconstruction error before and after isolating the mid-frequency components, using this as a cue for classification. Notably, to better align the frequency guidance with its subsequent impact on the reconstruction result, we innovatively integrate the reconstruction pipeline into the detection framework by using only the autoencoder (AE) of a latent DM (LDM), thus bypassing the complex denoising process. This end-to-end learning approach better aligns the DM with the detection task, enhancing the consistency of the mask refinement process.

In summary, the contributions of our work are threefold:
\begin{itemize}
\item We are the first to integrate frequency decomposition into reconstruction-based detection methods, identifying the image components that contribute to higher reconstruction errors.
\item We propose an innovative end-to-end learning approach that better aligns the classifier with the task of detecting images generated by DM.
\item Extensive visualizations validate our hypothesis regarding the frequency distribution of reconstruction errors, and comprehensive experiments demonstrate the effectiveness of our proposed method.
\end{itemize}

%% file: sec/2_related_works.tex
\section{Related Works}
\label{sec:related_works}

\begin{figure*}[ht]
    \centering
    \includegraphics[width=0.8\textwidth]{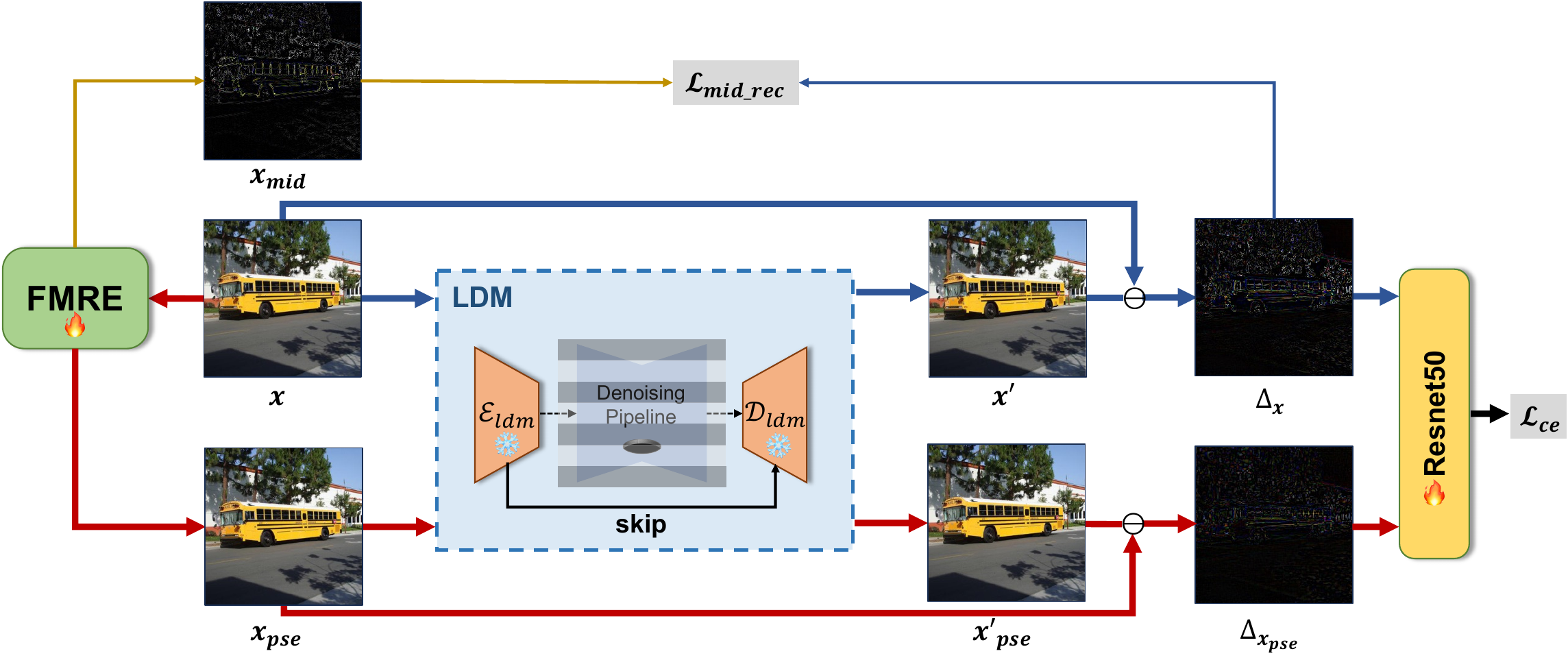}
    \caption{\textbf{The overview of FIRE.} We aim to \textbf{extract the frequency bands from the image that the diffusion model struggles to reconstruct}, i.e. information that is abundant in real images but lacking in generated ones, and then 
    \textbf{compare the reconstruction errors before and after the extraction to determine whether the image is real or generated.} The original image first undergoes reconstruction error computation using an LDM, where we substitute the AE of LDM for the reconstruction process to avoid introducing the denoising pipeline. To effectively extract the frequency band information that is difficult for the diffusion model to reconstruct, we propose the \textbf{F}requency \textbf{M}ask \textbf{RE}finement Module (FMRE). The reconstruction error is then computed for the pseudo-generated image with such information removed. Finally, the two reconstruction error maps are concatenated along the channel dimension and fed into the classifier.}
    \label{fig:method}
\end{figure*}

\begin{figure}[ht]
    \centering
    \includegraphics[width=\linewidth]{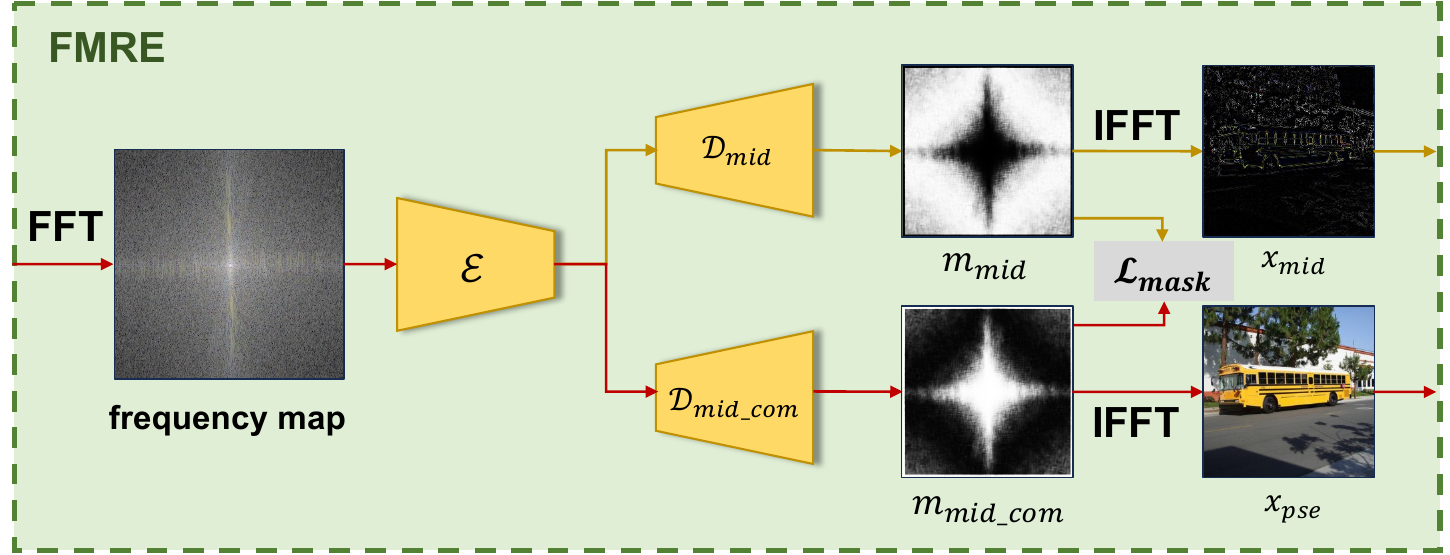}
    \caption{The architecture of our proposed FMRE, which consists of a shared encoder and two independent decoders.}
    \label{fig:fmre}
\end{figure}

\subsection{Generated Image Detection}
With advancements in AI-based image generation, numerous detection methods have emerged to tackle the challenges of increasingly realistic synthetic images. The primary approach utilizes neural networks to capture artifacts within images \cite{ba2024exposing,liu2023bifpro,huang2023implicit}. Zhong et al. \cite{zhong2024patchcraft} detect generated images by segmenting them into patches and examining inter-pixel correlations, while Tan et al. \cite{tan2024rethinking} introduce Neighboring Pixel Relationships, which identify generated content by analyzing local pixel distribution patterns during upsampling.

Additionally, multimodal methods leverage semantic information to enhance detection \cite{zou2024cross,yin2024improving}. Chang et al. \cite{chang2023antifakeprompt} and Jia et al. \cite{jia2024can} reformulate detection as a visual question answering task, combining vision-language models such as InstructBLIP \cite{dai2023instructblip} and ChatGPT \cite{achiam2023gpt} to improve performance on unseen data. Shao et al. \cite{shao2023detecting} propose HAMMER for detecting and attributing manipulated content by examining subtle interactions between image and text.

Frequency domain analysis has also proven effective \cite{miao2023f,tan2024frequency,woo2022add}, complementing spatial methods that struggle with certain artifacts. Dolhansky et al. \cite{doloriel2024frequency} use frequency-based masking to extract common features, while Li et al. \cite{li2024freqblender} classify frequency components into semantic, structural, and noise levels to locate regions with distinctive frequency distributions.

\subsection{Detection Based on Reconstruction Error}
The advent of DMs \cite{ho2020denoising} has spurred the development of specialized detectors to identify DM-generated images. Wang et al. \cite{wang2023dire} propose Diffusion Reconstruction Error (DIRE), which differentiates real from DM-generated images by measuring reconstruction error. Ma et al. \cite{ma2023exposing} enhance detection accuracy using multi-step error computation in their SeDID method. Ricker et al. \cite{ricker2024aeroblade} utilize autoencoder reconstruction error from latent DMs for a simple, training-free approach. Cazenavette et al. \cite{cazenavette2024fakeinversion} develop FakeInversion to detect images generated by unseen text-to-image DMs using text-conditioned inversion. Chen et al. \cite{chendrct} introduce Reconstruction Contrastive Learning to improve generalization by generating hard samples. Additionally, Luo et al. \cite{luo2024lare} propose LaRE2, leveraging Latent Reconstruction Error (LaRE) with an Error-Guided Feature Refinement module for more distinct error feature extraction.

Compared to existing methods, our approach is the first to explore the impact of frequency decomposition on reconstruction error. By exploring the hard-to-reconstruct frequency bands hidden within the image, we build a robust detector specifically targeting DM-generated images.

%% file: sec/3_method.tex
\section{Method}

In this section, we first provide foundational knowledge on DDPM and the principles behind reconstruction error-based detection methods. Then, we introduce our method named \textbf{F}requency-gu\textbf{I}ded \textbf{R}econstruction \textbf{E}rror (\textbf{FIRE}), which leverages the observation that DMs struggle to reconstruct mid-frequency information in real images. By comparing the differences between real and fake images after frequency-guided reconstruction, FIRE determines whether an image is generated by a DM. The overall framework of FIRE is illustrated in Figure~\ref{fig:method}.

\subsection{Diffusion Reconstruction Pipeline}
\label{sec:method_3_1}
Existing DMs can be summarized into two stages: the forward process and the reverse process. In the forward process, noise of increasing intensity is progressively added to the input image:
\begin{align}
    x_{t}=\sqrt{\bar{\alpha}_{t}} x_{0}+\sqrt{1-\bar{\alpha}_{t}} \epsilon,
\end{align}
where $\epsilon_{t} \sim \mathcal{N}(0, I)$ for $t = 0, \dots, T$. Here, $x_t$ represents the noisy image at time step $t$, and $x_0$ represents the initial image without noise. $\sqrt{\bar{\alpha}{t}}$ denotes the noise scaling factor at step $t$, and $\bar{\alpha}{t} = \prod_{s=1}^{t} \alpha_{s}$.

In the reverse process, the image is gradually denoised back to its clean state, primarily through the use of a parameterized neural network that predicts the denoised result at each time step $t$, as follows:
\begin{align}
    x_{t-1}=\sqrt{\alpha_{t-1}} \frac{x_{t}-\sqrt{1-\alpha_{t}} \epsilon_{\theta} ( x_{t}, t )} {\sqrt{\alpha_{t}}}+\sqrt{1-\alpha_{t-1}} \epsilon_{t},
\end{align}
where $\alpha_{t-1} = \frac{\bar{\alpha}{t-1}}{\bar{\alpha}{t}}$ for $t = T, \dots, 1$, and $\epsilon_{\theta}(x_{t}, t)$ is the predicted noise by the denoising neural network parameterized by $\theta$, typically implemented as a U-Net.

\subsection{FIRE}
As illustrated in Section~\ref{sec:intro_hypothesis}. We hypothesize that the mid-frequency components of real images are particularly difficult for DMs to reconstruct. Thus, if a real image has these mid-frequency components removed, its reconstruction error will be smaller than that of the unaltered image. In other words, by stripping away some of the inherent information that differentiates real images from generated ones, the modified real image becomes more similar to a generated image—we term this a \textbf{pseudo-generated image}. Intuitively, the difference between a real image and its pseudo-generated version will be greater than the difference between a generated image and its pseudo-generated version. Thus, this frequency-guided difference can serve as a cue for distinguishing real from generated images.

To this end, we propose FIRE, which distinguishes real images from DM-generated ones by comparing reconstruction error differences between an image and its pseudo-generated version. First, we apply FFT to image $x$ and center the low-frequency components. Then, a Frequency Mask Refinement module (FMRE) identifies mid-frequency regions that are difficult for the DM to reconstruct. The frequency mask is applied to obtain a band-pass filtered pseudo-generated image $x_{pse}$. Both the original and pseudo-generated images are then input to the DM for reconstruction.
\begin{align}
    x^{\prime}_{pse} &= \mathrm{R}(x_{pse}), \\ x^{\prime} &= \mathrm{R}(x), 
\end{align}
where $\mathrm{R}(\cdot)$ represents the reverse and reconstruction process of a DM. Notably, recent work \cite{ricker2024aeroblade} has shown that the denoising process is not critical for the image generation detection task, and satisfactory performance can be achieved by using only the AE of LDM to compute the reconstruction error. Following this insight, we use only the encoder and decoder of a LDM to replace the entire diffusion reconstruction pipeline:
\begin{align}
    \mathrm{R}=\mathcal{D}_{\mathrm{ldm}}(\mathcal{E}_{\mathrm{ldm}}(\cdot)),
\end{align}
where $\mathcal{E}_{\mathrm{ldm}}$ and $\mathcal{D}_{\mathrm{ldm}}$ denote the encoder and decoder of the LDM, respectively. This design reduces training costs, which has driven existing methods to use a two-step process: reconstructing the image with a DM and then separately training on the reconstruction. Our approach directly links the detection module with the AE, enabling end-to-end training that aligns the learned representations of the detector with the latent space of the DM.

Once we obtain the reconstructed images $x^{\prime}_{pse}$ and $x^{\prime}$, we compute their reconstruction error as follows:
\begin{align}
    \Delta_{x}&=\vert{x^{\prime}-x}\vert,\\ \Delta_{x_{pse}}&=\vert{x^{\prime}_{pse}-x_{pse}}\vert,
\end{align}
where $|\cdot|$ denotes the absolute difference. Finally, we concatenate the two reconstruction error maps and feed them into a binary classifier for final prediction:
\begin{align}
    y^{\prime}=\theta_{cls}(\mathrm{cat}(\Delta_{x},\Delta_{x_{pse}})),
\end{align}
where $\theta_{cls}$ represents the binary classifier, $\mathrm{cat}(\cdot, \cdot)$ denotes concatenation along the channel dimension, and $y^{\prime}$ is the predicted label.

\subsection{Frequency Mask Refinement Module}
Our frequency mask refinement module uses an encoder-decoder architecture. The encoder has four convolutional layers (3 × 3 kernel, stride 2, padding 1), and the decoder includes one convolutional layer (3 × 3 kernel, stride 2, padding 1) followed by a PixelShuffle operation \cite{shi2016real}. The encoder $\mathcal{E}$ encodes the frequency spectrum after FFT, and two decoders, $\mathcal{D}_{\mathrm{mid}}$ and $\mathcal{D}_{\mathrm{mid\_c}}$, generate two distinct masks: $m_{\mathrm{mid}}$ for mid-frequency regions and $m_{\mathrm{mid\_c}}$ for complementary high- and low-frequency regions.
\begin{align}
    m_{\mathrm{mid}}&=\mathcal{D}_{\mathrm{mid}}(\mathcal{E}(\mathcal{F}(x))),\\m_{\mathrm{mid\_c}}&=\mathcal{D}_{\mathrm{mid\_c}}(\mathcal{E}(\mathcal{F}(x))),
\end{align}
where $\mathcal{F}(\cdot)$ denotes the Fast Fourier Transform. The mask $m_{\mathrm{mid}}$ is used to isolate the mid-frequency information that is challenging for the DM to reconstruct, and $m_{\mathrm{mid\_c}}$ is used to produce the pseudo-generated image:
\begin{align}
    x_{mid}&=\mathcal{F}^{-1}(\mathcal{F}(x)\otimes m_{\mathrm{mid}}),\\
    x_{pse}&=\mathcal{F}^{-1}(\mathcal{F}(x)\otimes m_{\mathrm{mid\_c}}),
\end{align}
where $\mathcal{F}^{-1}(\cdot)$ denotes the Inverse Fast Fourier Transform, $\otimes$ denotes element-wise multiplication.
The overall architecture of the module is shown in Figure~\ref{fig:fmre}.

\subsection{Loss Function}
Given an image, we aim for the frequency mask refinement module to precisely locate the mid-frequency band that aligns with the reconstruction error map. Intuitively, we transform the extracted mid-frequency components back to the spatial domain using IFFT and need to align them with the reconstruction error of raw image using the following \textbf{mid-frequency reconstruction alignment loss}:
\begin{align}
    {{\cal L}_{\mathrm{mid\_rec}}=}\Vert x_{mid}-\Delta_{x}\Vert^{2}_{2}.
\end{align}

Additionally, based on our prior spectral analysis, we have a preliminary understanding of the difficult-to-reconstruct frequency distribution in real images. We want the mid-band and complementary masks, predicted by the frequency mask refinement module, to capture the desired frequency bands and remain complementary to some extent. To accelerate model convergence and ensure that the predicted masks align with our expectations, we predefine two ideal masks. Given a frequency spectrum $f \in \mathbb{R}^{1 \times 256 \times 256}$ with a center point at $o = (128, 128)$, the predefined mid-frequency mask is:
\begin{align}
    {M_{mid}(u,v)}&=\begin{cases} {{1}} & {{\mathrm{i f ~} \mathrm{d}((u,v),o ) \in [40,120]}}, \\ {{0}} & {{\mathrm{o t h e r w i s e}}} ,\\ \end{cases}\\M_{mid\_c}&=\textbf{1}-M_{mid},
\end{align}
where $d(\cdot, \cdot)$ represents the Euclidean distance between coordinates, and $\mathbf{1} \in {1}^{1 \times 256 \times 256}$. Therefore, we apply the following \textbf{mask refinement loss} to the learned masks:
\begin{align}
    \begin{split}{{\cal{L}}_{\mathrm{mask}}=}&\Vert m_{mid}-M_{mid}\Vert^{2}_{2}+\Vert m_{mid\_c}-M_{mid\_c}\Vert^{2}_{2}\\&+\Vert\textbf{1}-m_{mid}-m_{mid\_c}\Vert^{2}_{2},\end{split}
\end{align}
Finally, the \textbf{overall loss objective} of our method minimizes the combination of the mid-frequency reconstruction alignment error, the mask refinement loss, and the cross-entropy loss:
\begin{align}
    {\cal{L}}=\lambda_0 {\cal{L}}_{\mathrm{mid\_rec}}+\lambda_1 {\cal{L}}_{\mathrm{mask}}+\lambda_2 \cal L_{\mathrm{ce},}
\end{align}
where $\mathcal{L}_{\mathrm{ce}}$ is the cross-entropy loss for training the binary classifier, and we use ResNet-50 as the backbone network:
\begin{align}
    {\cal{L}}_{ce}=-[(y\log ( y^{\prime} )+( 1-y ) \log ( 1-y^{\prime}))],
\end{align}
where $y$ represents the true label, $y^{\prime}$ is the predicted label, and $\lambda_0, \lambda_1, \lambda_2$ are balancing factors for the three loss terms. In our experiments, $\lambda_0$ and $\lambda_1$ are set to $\frac{1}{5}$, and $\lambda_2$ is set to $\frac{3}{5}$.

%% file: sec/4_experiment.tex
\section{Experiment}
\label{sec:experiment}
In this section, we first describe the experimental details and then provide comprehensive experimental results to validate the effectiveness of FIRE.
\subsection{Setup}
\textbf{Baselines.}
We compare our method against a set of state-of-the-art detectors. All methods use the official open-source code for training and inference on our experimental datasets. CNNDet \cite{wang2020cnn} employs a CNN to detect images at the RGB level, revealing that generated images can be effectively identified by convolutional classifiers. AEROBLADE \cite{ricker2024aeroblade} adopts a training-free detection approach by calculating the reconstruction error after it passes through the AE of LDM. DIRE \cite{wang2023dire} explores image-level DDIM reconstruction error as detection cues. FakeInversion \cite{cazenavette2024fakeinversion} utilizes the latent noise map from the inversion process of Stable Diffusion, guided by prompts, along with the reconstructed image as additional input signals for detection.

\input{table/table0.tex}
\input{table/table1.tex}

\noindent\textbf{Datasets.}
We conduct experiments on two datasets, DiffusionForensics \cite{wang2023dire} and a self-collected dataset. \\
\noindent\textbf{DiffusionForensics} is a relatively simple open-source benchmark. We chose to use the LSUN-bedroom and ImageNet subsets for experiments. LSUN-bedroom subset collects bedroom images from LSUN-Bedroom \cite{yu2015lsun} and generates fake images using various DMs, including: ADM \cite{dhariwal2021diffusion}, DDPM \cite{ho2020denoising}, IDDPM \cite{nichol2021improved}, PNDM \cite{liu2022pseudo}, SD-v1 \cite{rombach2022high}, SD-v2 \cite{rombach2022high}, LDM \cite{rombach2022high}, VQ-Diffusion \cite{gu2022vector}, IF \cite{saharia2022photorealistic}, DALLE·2 \cite{ramesh2022hierarchical}, and Midjourney \footnotemark. The training set consists of 40,000 real images and 40,000 ADM-generated images. For each subset, we select 1,000 real and 1,000 generated images for testing. Additionally, the ImageNet subset is crafted to evaluate models in general scenarios. Specifically, they use ADM \cite{dhariwal2021diffusion} and SD-v1 \cite{rombach2022high} to generate images. We also use 40,000 real and 40,000 ADM-generated images for training, with 5,000 real and 5,000 generated images selected for testing across both ADM and SD-v1 subsets.
\footnotetext{\url{https://www.midjourney.com}}

\noindent\textbf{Self-collected dataset.} To assess the performance of our model in more realistic scenarios, we follow \cite{cazenavette2024fakeinversion} and form a new dataset using several novel DMs. For real images, we randomly sample 10,000 images from LAION \cite{schuhmann2022laion} for training, with 1,000 images for testing. For generated images, we use prompts from Midjourney and generate 10,000 images using several open-source text-to-image models for training, with 1,000 for testing. These models include: DALL·E 3 \cite{betker2023improving}, Kandinsky 3 \cite{arkhipkin2023kandinsky}, Midjourney\footnotemark[\value{footnote}], SDXL \cite{podell2023sdxl}, and Segmind Vega \cite{gupta2024progressive}.

\noindent\textbf{Implementation details.} For the data preprocessing, we apply a series of random augmentations, including flip, crop, color jitter, grayscale, cutout, noise, blur, jpeg, and rotate. Each image is resized to 256 $\times$ 256 along the shortest side. During training, the batch size is set to 16, and we use the Adam optimizer with an initial learning rate of 1e-4. We train for 100 epochs, and all experiments are conducted on a single NVIDIA A100 GPU. We adopt two widely used metrics for image generation detection: Area Under ROC (AUC) and accuracy (ACC), to evaluate the effectiveness of models. It is worth mentioning that we use the AE from SD-v1.5 \cite{runwayml2023sd} to compute reconstruction errors. Stable Diffusion uses a variational autoencoder (VAE) \cite{kingma2013auto} with Kullback-Leibler regularization.

\subsection{Comparison to Baselines}
\subsubsection{Performance on DiffusionForensics}
Table~\ref{tab:mainresults} shows the performance of different detection models under various generation methods on the LSUN-Bedroom and ImageNet datasets. We observe that existing detectors, such as CNNDetection, exhibit significant performance drops when dealing with unseen datasets and unseen diffusion methods. While DIRE and FakeInversion reach satisfactory performance, they still show slight drops in a few unseen domains. In comparison, our FIRE method achieves an average 100\% AUC and ACC across all tests.

\subsubsection{Performance on Newer Generation Models}
The above results demonstrate that the DiffusionForensics dataset is not a challenging task for our method. We then evaluate models on the five newer generation methods, as shown in Table~\ref{tab:second}. The results show that CNNDet, AEROBLADE, and DIRE exhibit obvious deficiencies in cross-model generalization. The result suggests a fundamental limitation of these methods when handling unseen generation models. In contrast, our proposed FIRE method achieved significant performance improvements in all test scenarios. Specifically, FIRE outperform the next-best method FakeInversion by 3.2\% AUC and 4.6\% ACC when trained on LAION + DALL·E 3, and by 7.6\% AUC and 7.3\% ACC on LAION + Kandinsky 3. These experiments reveal that FIRE strikes a better balance between performance and generalization.

\subsection{Ablation Study}
In this section, we conduct several ablation studies to evaluate the contribution of each component in the model and explore the impact of different predefined frequency masks.

\input{table/table2.tex}
\input{table/table3.tex}

\begin{figure}[ht]
    \centering
    \includegraphics[width=\linewidth]{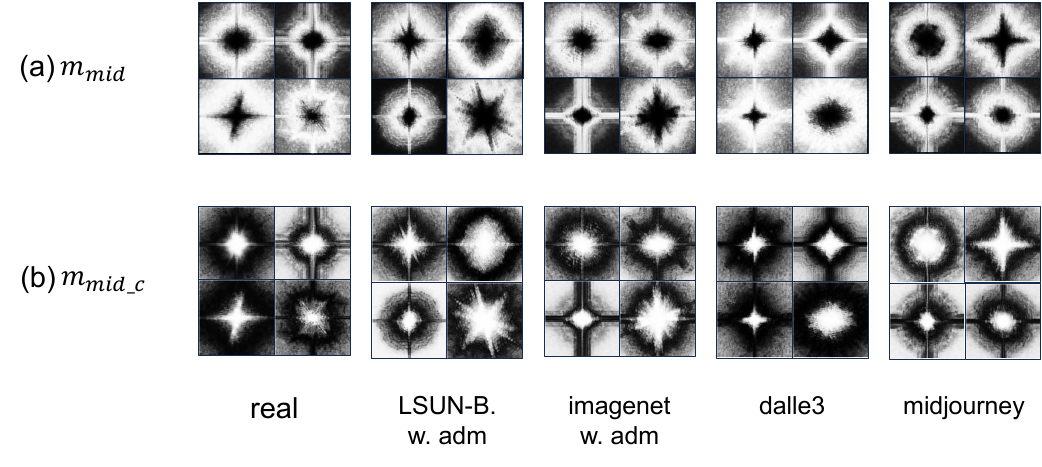}
    \caption{\textbf{Visualization of Filtered Frequency Maps in FMRE.} (a) highlights frequency bands the model focuses on, likely containing hard-to-reconstruct information. (b) shows the residual frequency map post-filtering, used for pseudo-generated images. The model primarily focuses on mid-band frequencies, supporting the hypothesis in Section~\ref{sec:method_3_1}. Additionally, (a) and (b) are complementary, indicating that (a) can be fully decoupled from (b).}
    \label{fig:vision_mask}
\end{figure}

\begin{figure}[ht]
    \centering
    \includegraphics[width=\linewidth]{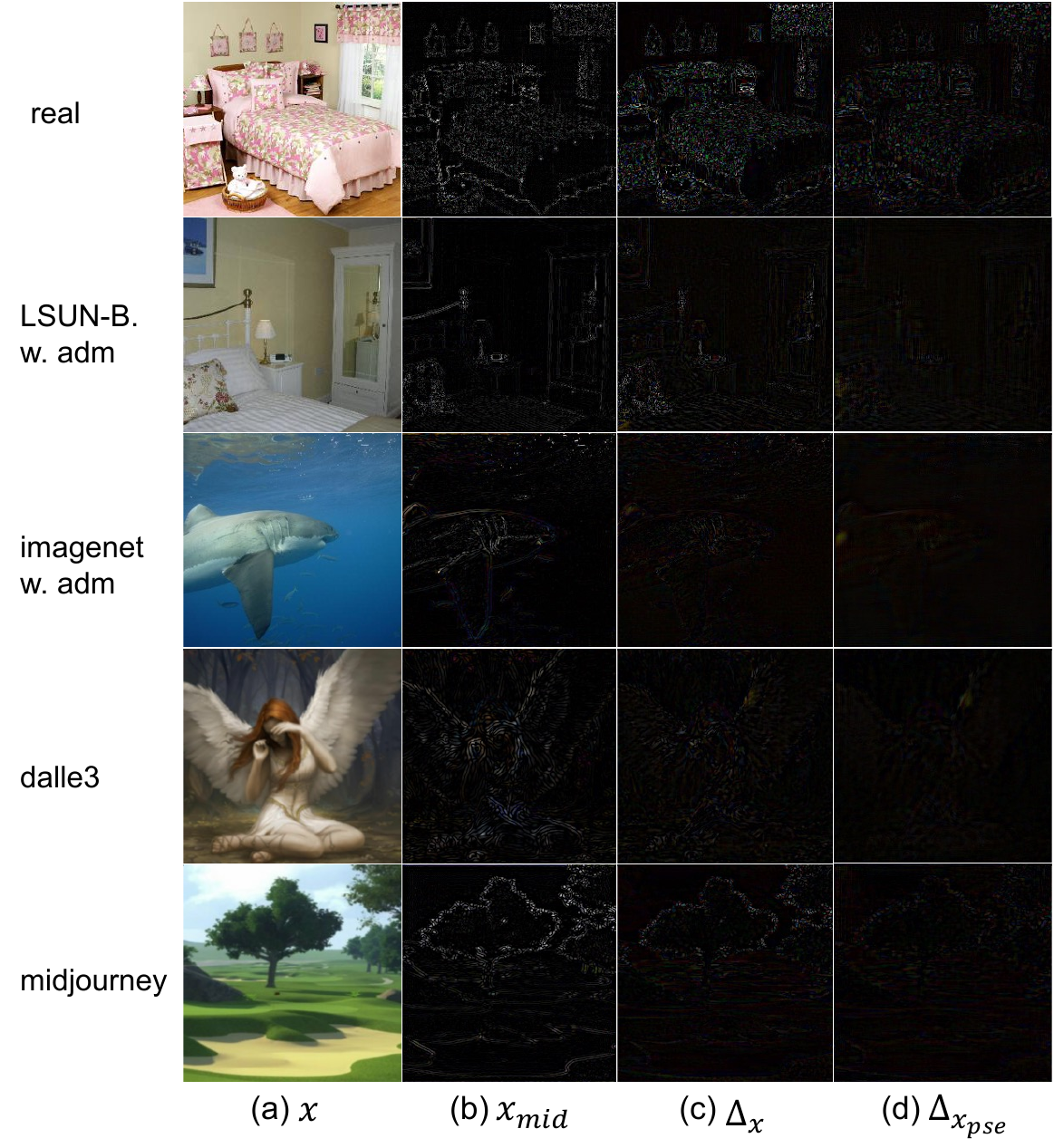}
    \caption{\textbf{Visualization of Mid-Band Frequency Information and Reconstruction Errors.} (a) shows original images. (b) displays images after applying IFFT to mid-band frequencies extracted by FMRE. (c) represents reconstruction errors of (a), while (d) shows those of pseudo-generated images. The reconstruction error for frequency-filtered pseudo-generated images is significantly lower than for real images, with real images showing a more pronounced change in error before and after filtering.}
    \label{fig:vision_rec} 
\end{figure}

\begin{figure}[ht]
    \centering
    \includegraphics[width=0.4\textwidth]{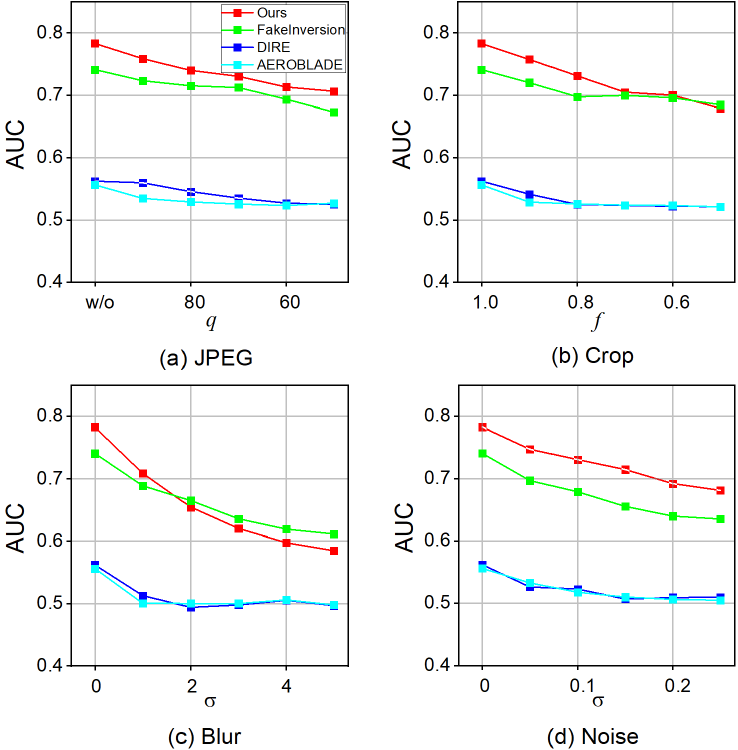}
    \caption{Detection performance of FIRE and baselines when handling perturbed images, measured in AUC.}
    \label{fig:perturb}
\end{figure}

\subsubsection{Influence of Predefined Frequency Mask}
To verify that our predefined frequency mask $M_{mid}$ indeed helps the model focus on the mid-band frequency that DMs struggle to reconstruct, we designed an ablation study. We use different preset frequency masks as shown in Figure~\ref{fig:fig1}. We then applied these different mask settings and train on LAION + DALL·E 3 for performance comparison, as shown in Table~\ref{tab:maskablation}. As indicated in the table, the experimental results confirm that the key information for effectively distinguishing real images is indeed hidden in the mid-frequency components of the image.

\subsubsection{Influence of Loss Terms}
In this section, we conduct ablation studies on each objective term of the model, with the experiment also conducted on LAION + DALL·E 3. The results are shown in Table~\ref{tab:lossablation}. We first ablate the ${\cal{L}}_{mask}$, meaning the model lacks the guidance of the predefined frequency mask and must spontaneously discover the frequency regions corresponding to the reconstruction error map. The results show that the model's performance drops under this setting, indicating that the predefined frequency mask helps the model better locate the frequency regions corresponding to the reconstruction error. Next, we ablate the ${\cal{L}}_{mid\_rec}$, resulting in a significant performance degradation. Since our method relies on the alignment between the reconstruction error and the extracted mid-frequency information, ${\cal{L}}_{mid\_rec}$ is crucial to the overall approach. Removing it reduces the method to a simpler approach similar to DIRE \cite{wang2023dire}. These analyses demonstrate that each objective term contributes to the overall performance.

\subsection{Robustness to Unseen Perturbations}
In real-world scenarios, the images to be detected are often post-processed, such as through quality compression or cropping. In this section, we evaluate the robustness of our method to various perturbations. Following previous works \cite{frank2020leveraging,yu2020responsible}, we use the LAION + DALL·E 3 dataset and apply JPEG compression (quality factor q), center cropping (crop factor f and subsequent resizing), Gaussian blur, and Gaussian noise (both with standard deviation $\sigma$). The results are shown in Figure~\ref{fig:perturb}. In the figure, we report the performance of several SOTA methods and our approach under varying levels of image perturbation. We find that our method outperforms other detectors across most metrics. Only under severe blur conditions does our model perform worse than FakeInversion, likely because blurring destroys much of the high-frequency information in the image, reducing the gap between real and generated images in terms of reconstruction error, making it harder for the model to distinguish. In future work, we plan to design a more robust frequency refinement scheme to improve the robustness of FIRE.

\subsection{Visual Analysis}
\label{sec:vision}
To better understand our model’s behavior, we visualize the frequency maps decoded by two decoders of FMRE in Figure~\ref{fig:vision_mask}. The results show that the model mainly focuses on mid-band frequencies, supporting our hypothesis that these frequencies are crucial for distinguishing real from generated images.

In Figure~\ref{fig:vision_rec}, we further visualize mid-band frequency information and the corresponding reconstruction errors. Results indicate that reconstruction errors for frequency-filtered pseudo-generated images are significantly lower, as filtering reduces complex textures in the mid-band. This change is more pronounced in real images, suggesting that our model effectively captures frequency information that DMs struggle to reconstruct. Additional visualizations are in Section~\ref{sec:more_visualization}.

%% file: table/table0.tex
\begin{table*}[ht]
    \small
    \centering
    \setlength\tabcolsep{2pt}
    \resizebox{1.\linewidth}{!}{
    \begin{tabular}{clllccccclccccclllll}
        \cline{1-15}
        \multicolumn{1}{c|}{\multirow{2}{*}{Eval set}} & \multicolumn{2}{c}{\makecell{Train set w. \\ Gen. method}} & & \multicolumn{5}{c}{LSUN-B. w ADM} &  & \multicolumn{5}{c}{Imagenet w ADM} &  &  &  &  &  \\ \cline{2-3} \cline{5-9} \cline{11-15}
        \multicolumn{1}{c|}{} & \multicolumn{1}{c|}{\makecell{Gen. \\ method}} & Model & & CNNDet\cite{wang2020cnn} & AEROBLADE\cite{ricker2024aeroblade} & DIRE\cite{wang2023dire} & FakeInversion\cite{cazenavette2024fakeinversion} & FIRE(ours) &  & CNNDet & AEROBLADE & DIRE & FakeInversion & FIRE(ours) &  &  &  &  &  \\ \cline{1-15}
        \multicolumn{1}{c|}{\multirow{2}{*}{Imagenet\cite{deng2009imagenet}}} & \multicolumn{2}{l}{ADM\cite{dhariwal2021diffusion}} &  & 73.4/57.6 & 99.1/98.3 & 99.4/96.4 & \textbf{100.0}/99.8 & \textbf{100.0/100.0} &  & 99.6/99.3 & \textbf{100.0/100.0} & \textbf{100.0/100.0} & \textbf{100.0/100.0} & \textbf{100.0/100.0} &  &  &  &  &  \\
        \multicolumn{1}{c|}{} & \multicolumn{2}{l}{SD-v1\cite{rombach2022high}} &  & 67.3/53.4 & 98.7/97.4 & 98.3/95.2 & 99.7/97.6 & \textbf{100.0/100.0} &  & 87.2/84.7 & 99.7/98.3 & \textbf{100.0}/99.6 & \textbf{100.0/100.0} & \textbf{100.0/100.0} &  &  &  &  &  \\ \cline{1-15}
        \multicolumn{1}{c|}{\multirow{11}{*}{LSUN-B.\cite{yu2015lsun}}} & \multicolumn{2}{l}{ADM} &  & 96.8/93.7 & \textbf{100.0/100.0} & \textbf{100.0/100.0} & \textbf{100.0/100.0} & \textbf{100.0/100.0} &  & 76.0/54.2 & 98.4/97.5 & \textbf{100.0}/99.7 & \textbf{100.0/100.0} & \textbf{100.0/100.0} &  &  &  &  &  \\
        \multicolumn{1}{c|}{} & \multicolumn{2}{l}{DDPM\cite{ho2020denoising}} &  & 75.3/55.1 & 98.9/97.8 & \textbf{100.0/100.0} & \textbf{100.0/100.0} & \textbf{100.0}/99.8 &  & 68.4/45.1 & 99.1/98.2 & \textbf{100.0}/97.7 & \textbf{100.0/100.0} & \textbf{100.0/100.0} &  &  &  &  &  \\
        \multicolumn{1}{c|}{} & \multicolumn{2}{l}{IDDPM\cite{nichol2021improved}} &  & 76.2/49.5 & 99.7/98.2 & \textbf{100.0/100.0} & 99.8/98.4 & \textbf{100.0/100.0} &  & 77.6/53.8 & 99.5/98.7 & \textbf{100.0/100.0} & 98.6/97.9 & \textbf{100.0/100.0} &  &  &  &  &  \\
        \multicolumn{1}{c|}{} & \multicolumn{2}{l}{PNDM\cite{liu2022pseudo}} &  & 81.1/49.0 & 99.2/97.9 & 99.7/88.6 & \textbf{100.0}/99.7 & \textbf{100.0/100.0} &  & 73.2/38.7 & 97.9/97.4 & \textbf{100.0/100.0} & \textbf{100.0/100.0} & \textbf{100.0/100.0} &  &  &  &  &  \\
        \multicolumn{1}{c|}{} & \multicolumn{2}{l}{SD-v1\cite{rombach2022high}} &  & 69.7/49.7 & 99.6/98.4 & \textbf{100.0}/99.8 & \textbf{100.0/100.0} & \textbf{100.0/100.0} &  & 66.8/54.3 & 99.3/98.1 & \textbf{100.0/100.0} & \textbf{100.0/100.0} & \textbf{100.0/100.0} &  &  &  &  &  \\
        \multicolumn{1}{c|}{} & \multicolumn{2}{l}{SD-v2\cite{rombach2022high}} &  & 78.7/50.4 & 98.3/97.5 & \textbf{100.0/100.0} & \textbf{100.0/100.0} & \textbf{100.0/100.0} &  & 82.4/63.1 & 99.0/97.8 & \textbf{100.0/100.0} & \textbf{100.0}/99.9 & \textbf{100.0/100.0} &  &  &  &  &  \\
        \multicolumn{1}{c|}{} & \multicolumn{2}{l}{LDM\cite{rombach2022high}} &  & 59.4/49.2 & 99.8/98.5 & \textbf{100.0/100.0} & \textbf{100.0}/99.8 & \textbf{100.0/100.0} &  & 60.8/47.5 & 98.6/97.7 & \textbf{100.0}/98.6 & \textbf{100.0/100.0} & \textbf{100.0/100.0} &  &  &  &  &  \\
        \multicolumn{1}{c|}{} & \multicolumn{2}{l}{VQD\cite{gu2022vector}} &  & 71.5/51.9 & 98.5/97.2 & 100.0/99.8 & \textbf{100.0/100.0} & \textbf{100.0/100.0} &  & 72.1/50.2 & 99.8/98.4 & \textbf{100.0/100.0} & \textbf{100.0}/99.7 & \textbf{100.0/100.0} &  &  &  &  &  \\
        \multicolumn{1}{c|}{} & \multicolumn{2}{l}{IF\cite{saharia2022photorealistic}} &  & 78.0/50.3 & 99.3/97.6 & \textbf{100.0/100.0} & \textbf{100.0}/99.9 & \textbf{100.0/100.0} &  & 75.6/49.8 & 98.9/97.3 & \textbf{100.0/100.0} & \textbf{100.0}/99.9 & \textbf{100.0/100.0} &  &  &  &  &  \\
        \multicolumn{1}{c|}{} & \multicolumn{2}{l}{DALLE·2\cite{ramesh2022hierarchical}} &  & 78.3/67.1 & 99.0/98.1 & \textbf{100.0}/98.2 & \textbf{100.0/100.0} & \textbf{100.0/100.0} &  & 77.1/57.3 & 99.2/98.5 & \textbf{100.0}/99.9 & \textbf{100.0/100.0} & \textbf{100.0/100.0} &  &  &  &  &  \\
        \multicolumn{1}{c|}{} & \multicolumn{2}{l}{Mid.} &  & 86.1/74.6 & 97.9/97.3 & \textbf{100.0/100.0} & \textbf{100.0}/99.8 & \textbf{100.0/100.0} &  & 87.9/73.2 & 98.7/97.9 & \textbf{100.0/100.0} & \textbf{100.0}/99.7 & \textbf{100.0/100.0} &  &  &  &  &  \\ \cline{1-15}
        \multicolumn{3}{l}{Average} &  & 76.3/57.8 & 99.08/98.0 & 99.8/98.3 & \textbf{100.0}/99.6 & \textbf{100.0/100.0} &  & 77.3/59.3 & 99.1/98.1 & \textbf{100.0}/99.7 & 99.9/99.7 & \textbf{100.0/100.0} &  &  &  &  &  \\ \cline{1-15}
        \end{tabular}}
\caption{\textbf{Comparisons of our FIRE and other competitive state-of-the-art detectors.} We evaluate models on the offical DiffusionForensics dataset \cite{wang2023dire}. All the modles are re-trained with the official codes. We report AUC (\%) and ACC (\%) (AUC/ACC in the table).}
\label{tab:mainresults}
\end{table*}

%% file: table/table1.tex
\begin{table*}[ht]
    \small
    \centering
    \setlength\tabcolsep{2pt}
    \resizebox{1.\linewidth}{!}{
    \begin{tabular}{lllccccclccccc}
    \hline
    \multicolumn{2}{c}{Train set} &  & \multicolumn{5}{c}{{\color[HTML]{000000} LAION\cite{schuhmann2022laion} + DALLE·3}} &  & \multicolumn{5}{c}{LAION + Kan.3} \\ \cline{1-2} \cline{4-8} \cline{10-14} 
    \multicolumn{1}{l|}{Eval set} & Model &  & CNNDet & AEROBLADE & DIRE & FakeInversion & FIRE(ours) &  & CNNDet & AEROBLADE & DIRE & FakeInversion & FIRE(ours) \\ \hline
    \multicolumn{2}{l}{DALLE·3\cite{betker2023improving}} &  & 52.1/49.8 & 55.6/52.7 & 56.2/53.9 & 74.1/65.9 & \textbf{78.3/72.6} &  & 44.8/48.9 & 51.2/49.7 & 45.4/47.7 & 64.5/61.2 & \textbf{76.6/72.0} \\ \hline
    \multicolumn{2}{l}{Kan.3\cite{arkhipkin2023kandinsky}} &  & 46.3/50.0 & 53.5/51.0 & 52.4/49.6 & 68.3/62.7 & \textbf{74.5/68.7} &  & 51.5/47.6 & 56.6/53.8 & 55.2/52.1 & 78.7/73.8 & \textbf{86.5/79.9} \\ \hline
    \multicolumn{2}{l}{Mid.} &  & 50.6/50.0 & 47.6/50.2 & 53.1/50.2 & 59.6/54.5 & \textbf{66.2/64.7} &  & 45.6/50.0 & 48.6/49.9 & 52.1/49.8 & 65.0/58.2 & \textbf{71.4/66.9} \\ \hline
    \multicolumn{2}{l}{SDXL\cite{podell2023sdxl}} &  & 48.4/50.1 & 53.3/49.6 & 53.2/52.6 & \textbf{72.1/69.4} & 70.2/67.2 &  & 47.8/50.0 & 53.2/49.5 & 51.2/49.6 & 70.7/66.4 & \textbf{79.8/73.7} \\ \hline
    \multicolumn{2}{l}{Vega\cite{gupta2024progressive}} &  & 50.0/50.0 & 52.8/49.5 & 47.2/47.0 & 69.7/65.1 & \textbf{70.6/67.2} &  & 47.3/50.0 & 51.3/50.0 & 48.3/49.9 & 73.5/68.6 & \textbf{76.2/71.9} \\ \hline
    \multicolumn{2}{l}{Average} &  & 49.5/50.0 & 52.6/50.6 & 52.4/50.7 & 68.8/63.5 & \textbf{72.0/68.1} &  & 47.7/49.3 & 52.2/50.6 & 50.4/49.8 & 70.5/65.6 & \textbf{78.1/72.9} \\ \hline
    \end{tabular}}
    \caption{\textbf{Performance comparison of FIRE and baselines on updated generation methods.} Each model is loaded with weights pre-trained on the ImageNet subset of DiffusionForensics, then fine-tuned and tested on specific datasets. We report AUC (\%) and ACC (\%) (AUC/ACC in the table).}
    \label{tab:second}
\end{table*}

%% file: table/table2.tex
\begin{table}[ht]
    \resizebox{1.\linewidth}{!}{
        \begin{tabular}{cccccc}
            \hline
            Mask & DALLE·3 & Kan.3 & Mid. & SDXL & Vega \\ \hline
            \#1 & 58.5 & 57.3 & 55.8 & 62.1 & 55.2 \\ \hline
            \#2 & 73.4 & 68.4 & 61.6 & 68.5 & 62.7 \\ \hline
            \#3 & 62.7 & 56.9 & 54.4 & 66.8 & 54.3 \\ \hline
            \#4 & 76.5 & 72.2 & 65.8 & 69.4 & 68.9 \\ \hline
            \#5 (Ours) & \textbf{78.3} & \textbf{74.5} & \textbf{66.2} & \textbf{70.2} & \textbf{70.6} \\ \hline
        \end{tabular}}
    \caption{Detection performance using different preset frequency masks, measured in AUC.}
    \label{tab:maskablation}
\end{table}

%% file: table/table3.tex
\begin{table}[ht]
    \fontsize{11}{11}
    \resizebox{1.\linewidth}{!}{
    \begin{tabular}{ccccccc}
    \hline
    ${\cal{L}}_{\mathrm{mask}}$ & ${\cal{L}}_{\mathrm{mid\_rec}}$ & DALLE·3 & Kan.3 & Mid. & SDXL & Vega \\ \hline
     & \checkmark & 68.3 & 63.1 & 62.6 & 67.7 & 65.4 \\ \hline
     \checkmark &  & 54.3 & 55.6 & 53.4 & 55.3 & 51.3 \\ \hline
        \checkmark & \checkmark & \textbf{78.3} & \textbf{74.5} & \textbf{66.2} & \textbf{70.2} & \textbf{70.6} \\ \hline
    \end{tabular}}
    \caption{Ablation study on objective terms, measured in AUC.}
    \label{tab:lossablation}
\end{table}

%% file: sec/5_conclusion.tex
\section{Conclusion}
\label{sec:conclusion}
In this paper, we first analyze that real images contain inherent information that current DMs struggle to reconstruct, particularly concentrated in the mid-band frequency regions. Based on this observation, we propose a frequency-guided reconstruction error detection method, FIRE. FIRE captures such mid-band frequency information and removes it, detecting generated images by comparing the changes in reconstruction error before and after the removal. Unlike existing reconstruction-based approaches that rely on separate steps for calculating reconstruction errors and training classifiers, FIRE enables end-to-end learning, aligning the feature space for artifacts generation and detection. Extensive experiments demonstrate that FIRE outperforms state-of-the-art baselines, achieving superior performance both on standard datasets and under challenging conditions with perturbed images. We believe that FIRE offers a promising direction for improving the robustness and generalization for DM-generated image detection.

%% file: sec/X_suppl.tex
\clearpage
\setcounter{page}{1}
\maketitlesupplementary

\section{More Visualization}
\label{sec:more_visualization}
\begin{figure*}[ht]
    \centering
    \includegraphics[width=0.8\linewidth]{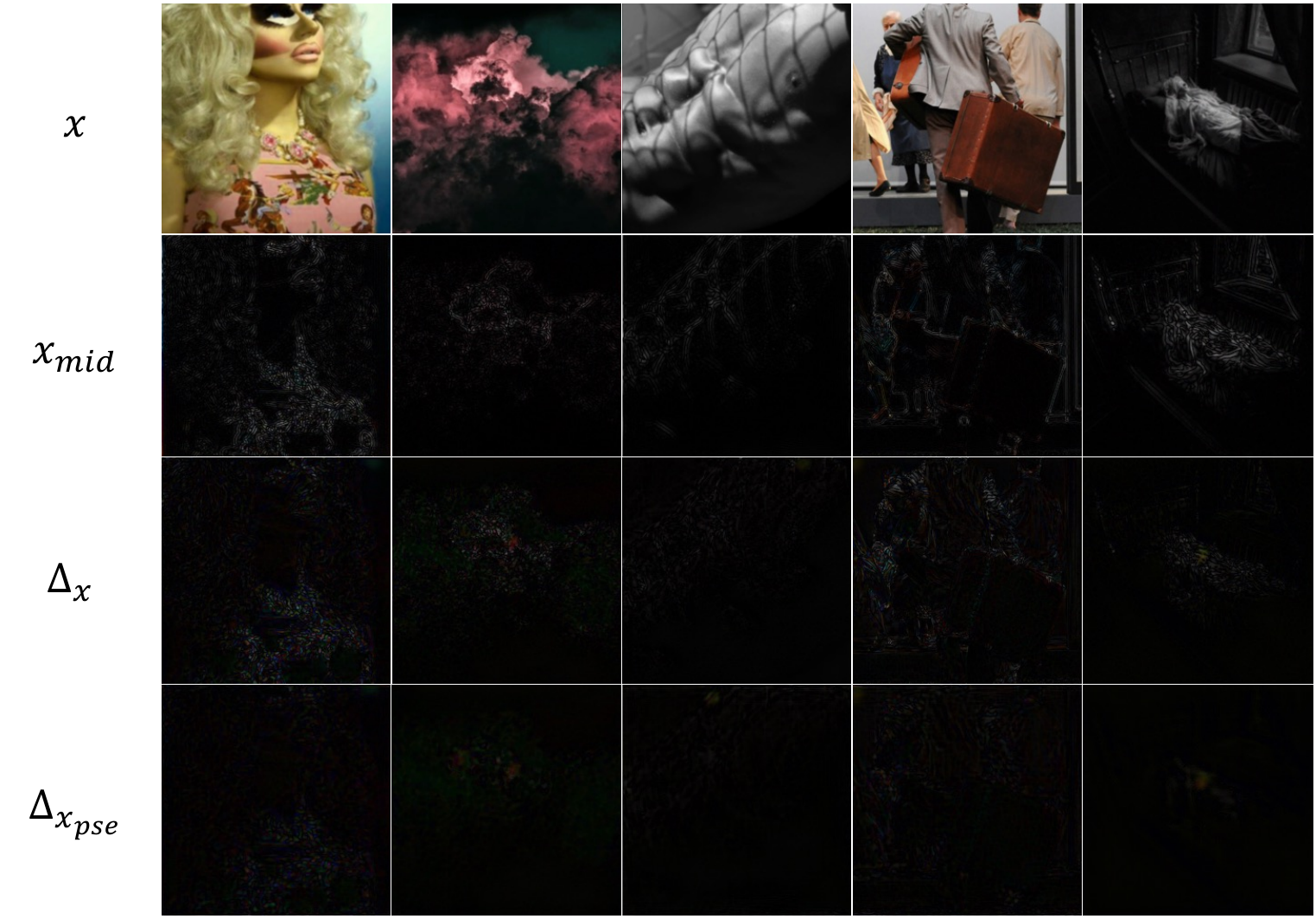}
    \caption{Visualization of mid-band frequency information and corresponding reconstruction errors on \textbf{real}. Random augmentations are applied.}
    \label{fig:x_0}
\end{figure*}

\begin{figure*}[ht]
    \centering
    \includegraphics[width=0.8\linewidth]{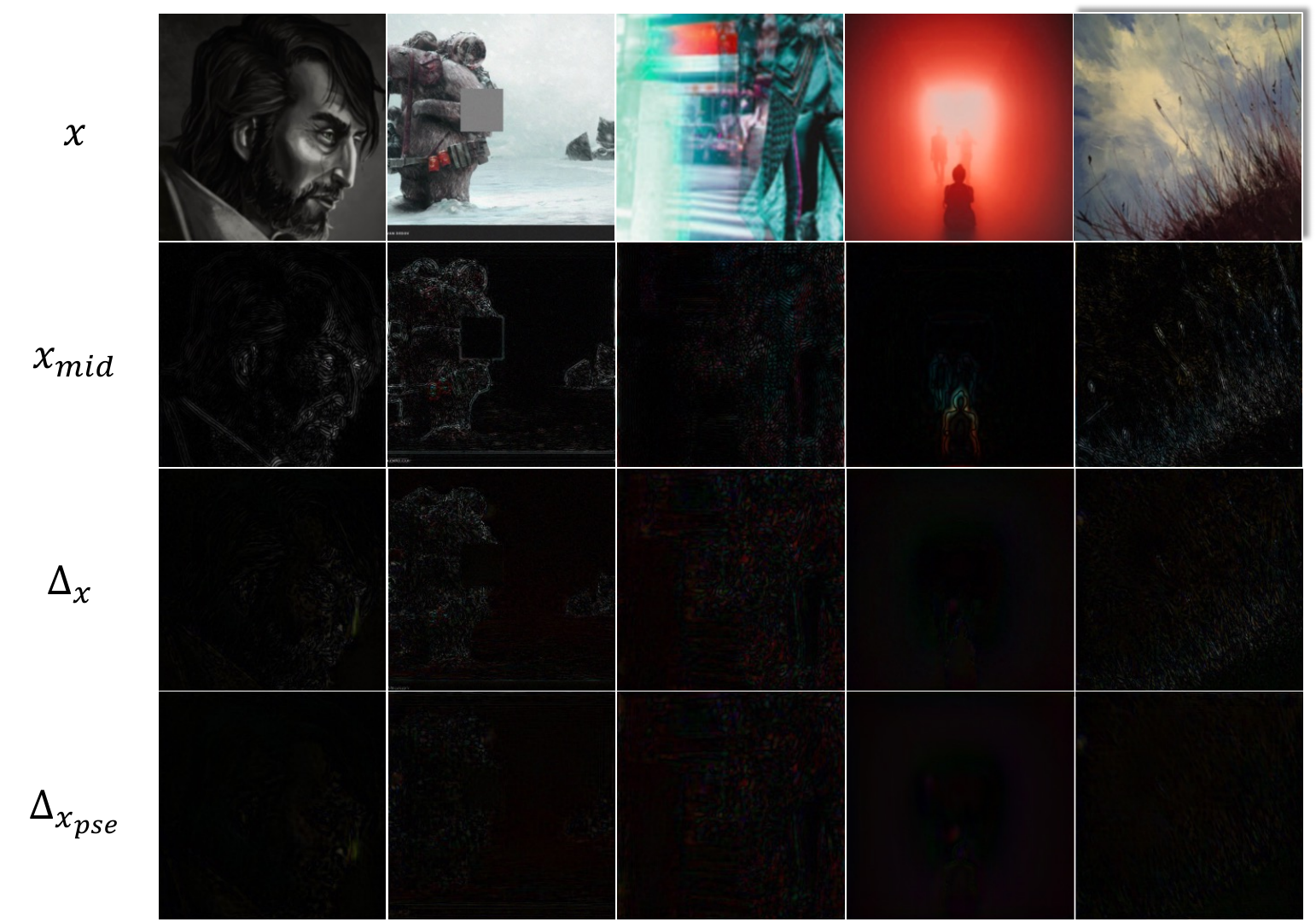}
    \caption{Visualization of mid-band frequency information and corresponding reconstruction errors on \textbf{midjounery}. Random augmentations are applied.}
    \label{fig:x_1}
\end{figure*}

\begin{figure*}[ht]
    \centering
    \includegraphics[width=0.8\linewidth]{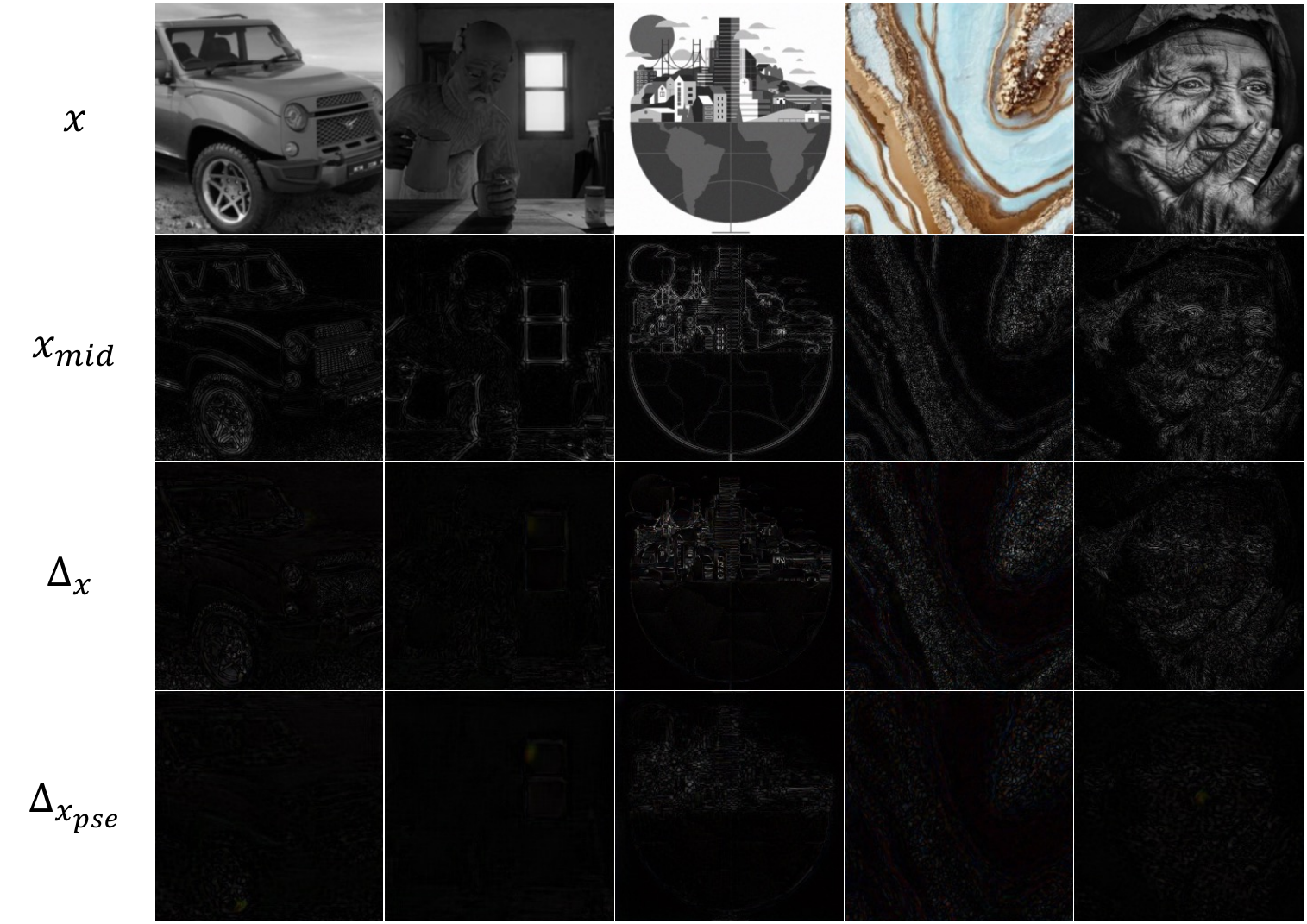}
    \caption{Visualization of mid-band frequency information and corresponding reconstruction errors on \textbf{Kandisary 3} \cite{arkhipkin2023kandinsky}. Random augmentations are applied.}
    \label{fig:x_2}
\end{figure*}

\begin{figure*}[ht]
    \centering
    \includegraphics[width=0.8\linewidth]{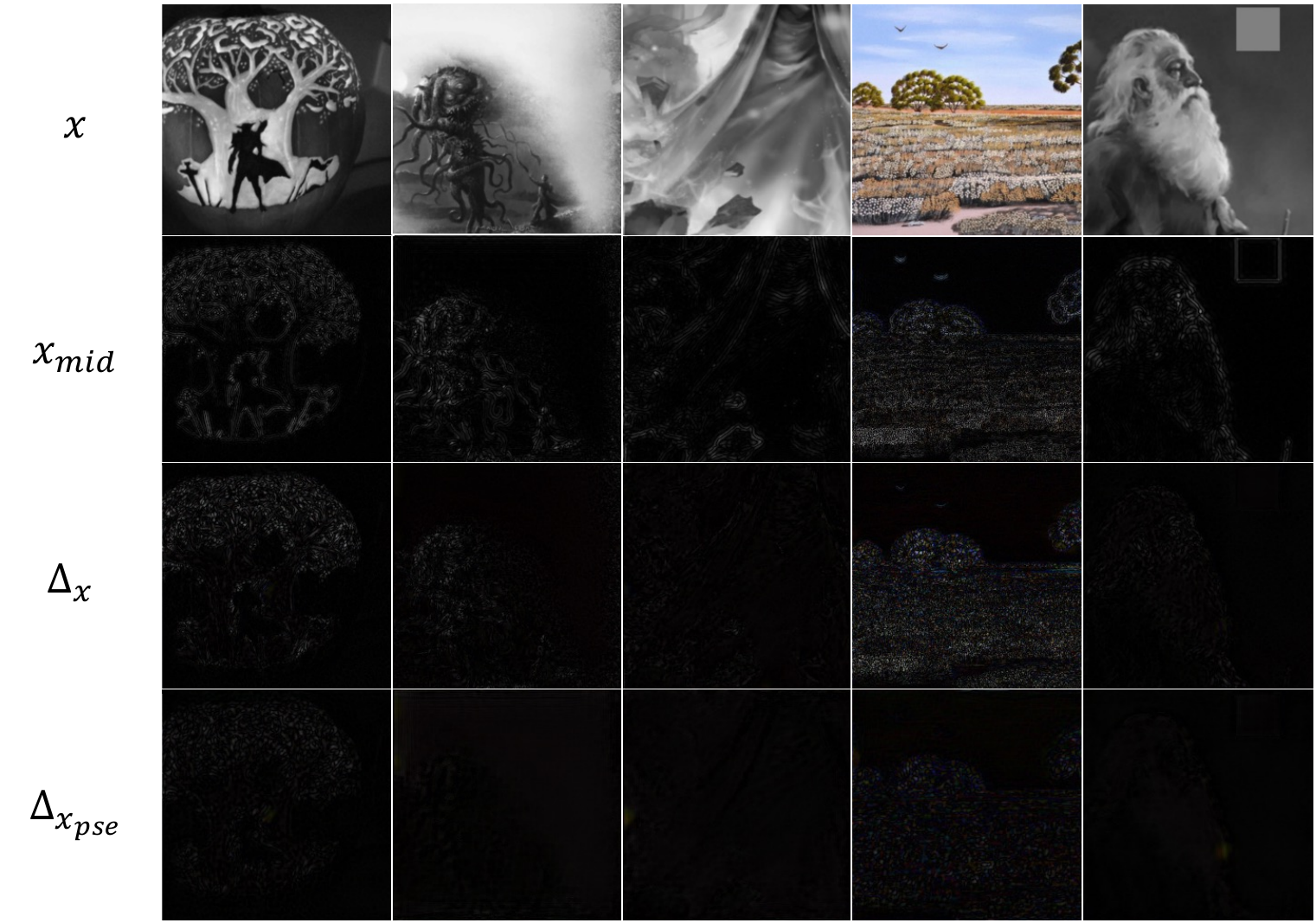}
    \caption{Visualization of mid-band frequency information and corresponding reconstruction errors on \textbf{DALL·E 3} \cite{betker2023improving}. Random augmentations are applied.}
    \label{fig:x_3}
\end{figure*}

\begin{figure*}[ht]
    \centering
    \includegraphics[width=0.8\linewidth]{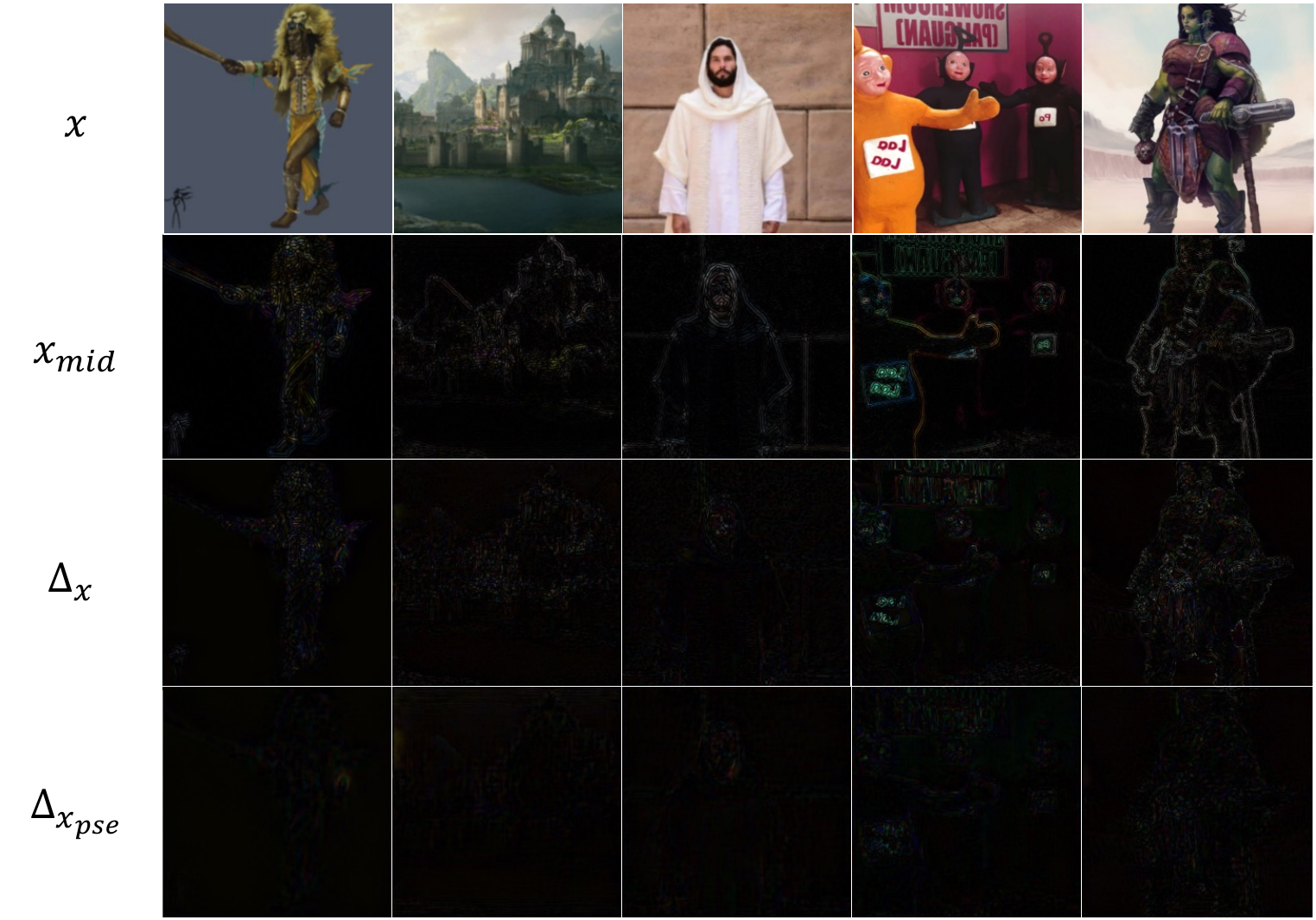}
    \caption{Visualization of mid-band frequency information and corresponding reconstruction errors on \textbf{SDXL} \cite{podell2023sdxl}. Random augmentations are applied.}
    \label{fig:x_4}
\end{figure*}

\begin{figure*}[ht]
    \centering
    \includegraphics[width=0.8\linewidth]{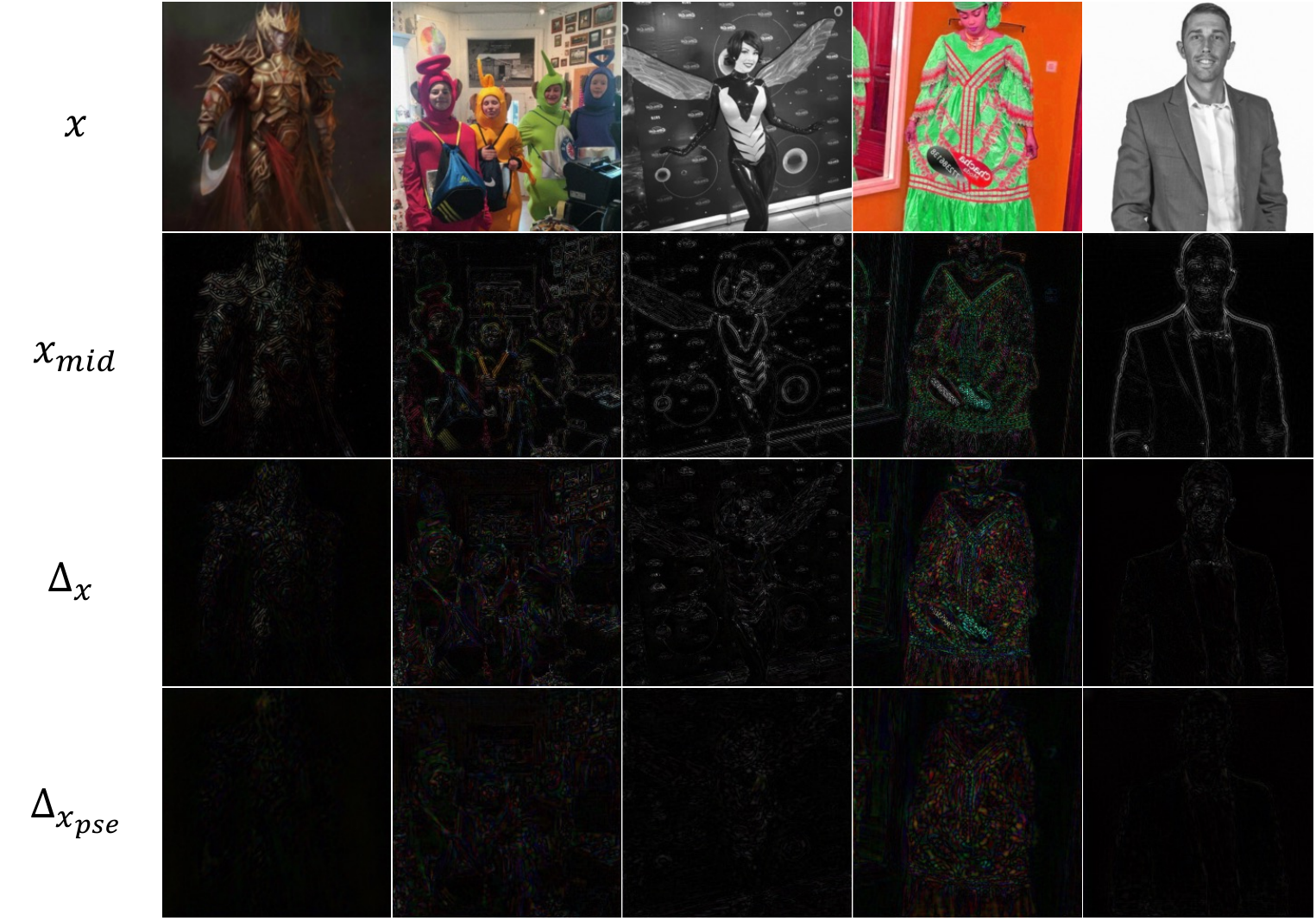}
    \caption{Visualization of mid-band frequency information and corresponding reconstruction errors on \textbf{Segmind Vega} \cite{gupta2024progressive}. Random augmentations are applied.}
    \label{fig:x_5}
\end{figure*}

We visualize more examples of mid-band frequency information extracted by FMRE and reconstruction errors on real, midjourney, Kandinsky 3 \cite{arkhipkin2023kandinsky}, DALL·E 3 \cite{betker2023improving}, SDXL\cite{podell2023sdxl} and Segmind Vega \cite{gupta2024progressive} in Figures~\ref{fig:x_0}-\ref{fig:x_5}. 

